\documentclass[11pt]{article}

\usepackage[margin=1in]{geometry}
\usepackage{amsmath,amssymb,amsfonts}
\usepackage{graphicx}
\usepackage{cite}
\usepackage{hyperref}
\usepackage{xcolor}
\usepackage{algorithm}
\usepackage{algorithmic}
\usepackage{lmodern}
\usepackage[T1]{fontenc}

\title{Fusing Semantic, Lexical, and Domain Perspectives for Recipe Similarity Estimation}

\author{
Denica Kjorvezir$^{1}$, Danilo Najkov$^{2}$, Eva Valenci\v{c}$^{1}$, \\
Erika Jesenko$^{3}$, Barbara Koro\u{s}i\'{c} Seljak$^{1}$, Tome Eftimov$^{1}$, Riste Stojanov$^{2}$ \\
$^{1}$Jo\v{z}ef Stefan Institute, Ljubljana, Slovenia \\
$^{2}$S.Cyril and Methodius University, Skopje, N. Macedonia \\
$^{3}$Biotechnical Faculty, University of Ljubljana, Slovenia \\
\texttt{\{dkorvezir, eva.valencic, barbara.korousic, tome.eftimov\}@ijs.si} \\
\texttt{\{danilo.najkov, riste.stojanov\}@finki.ukim.mk} \\
\texttt{erika.jesenko@bf.uni-lj.si}
}
\date{}
\begin{document}

\maketitle

\begin{abstract}
This research focuses on developing advanced methods for assessing similarity between recipes by combining different sources of information and analytical approaches. We explore the semantic, lexical, and domain similarity of food recipes, evaluated through the analysis of ingredients, preparation methods, and nutritional attributes. A web-based interface was developed to allow domain experts to validate the combined similarity results. After evaluating 318 recipe pairs, experts agreed on 255 (80\%). The evaluation of expert assessments enables the estimation of which similarity aspects---lexical, semantic, or nutritional---are most influential in expert decision-making. The application of these methods has broad implications in the food industry and supports the development of personalized diets, nutrition recommendations, and automated recipe generation systems.
\end{abstract}

\section{Introduction}

Scientific exploration of similar recipes transcends the boundaries of taste and aesthetics. It encompasses flavor chemistry, nutritional analysis, cultural insights, and the dynamic interplay between ingredients~\cite{1,2,3}. Studying recipes that share similarities enables food scientists to delve into the intricate realm of flavor chemistry analysis. Ingredients interact at a molecular level during cooking, leading to complex chemical reactions that influence taste, aroma, and texture~\cite{2,3}. By examining similar recipes, scientists can identify the molecular compounds responsible for specific flavor profiles. For instance, the combination of ingredients such as herbs, spices, and protein sources may result in synergistic or antagonistic interactions that ultimately shape a dish's sensory experience~\cite{1}. Such insights contribute to a deeper understanding of how flavors are developed and how these interactions can be manipulated to foster culinary creativity.

The scientific examination of similar recipes provides valuable insight into the nutritional composition of diverse culinary creations~\cite{4,5,6}. By comparing variations of a recipe, researchers can discern the nutritional implications of ingredient substitutions, omissions, or additions. This analytical approach is vital for understanding how different ingredient choices impact the macro- and micro-nutrient composition of a dish. For instance, replacing high-fat ingredients with healthier alternatives can reduce caloric content without compromising taste or texture. This nutritional analysis can contribute to the development of balanced and health-conscious recipes. By consistently selecting recipes that align with health goals, individuals can make meaningful contributions to their long-term health and quality of life.

Similar recipes also serve as a valuable dataset for understanding evolving consumer preferences and culinary trends~\cite{7,8,9}. Analyzing variations of popular dishes enables researchers to track shifts in ingredient choices, dietary preferences, and cultural influences. This information is invaluable for food companies, restaurants, and culinary entrepreneurs seeking to align their offerings with current consumer demands.

The analysis of similar recipes also aligns with the emerging field of data-driven analysis. Using computational intelligence models and machine learning (ML) algorithms, researchers can process vast recipe databases to predict ingredient pairings and suggest novel combinations. This integration of science and technology pushes the boundaries of food science, enhancing the efficiency and creativity of recipe development.

\section{Related Work}

\subsection{Lexical Similarity}

The comparison and identification of similar recipes has been an active area of research, particularly in the context of recipe recommendation systems. Freyne~\cite{10} proposed one of the early methods, recommending new recipes based on shared ingredients and user ratings. Although effective, this approach relies heavily on individual user preference data.

Kuo et al.~\cite{11} extended ingredient-based recommendation by proposing intelligent menu planning, generating sets of recipes that optimize ingredient coverage. This highlights the importance of ingredient similarity in recommending multiple complementary dishes.

Chhipa et al.~\cite{12} proposed a recipe recommendation system using TF-IDF vector representations of ingredients and textual descriptions. Cosine similarity was applied to compute similarity scores between recipes, providing an effective approach for identifying related dishes based on lexical content.

Wang et al.~\cite{9} introduced a substructure similarity measurement for Chinese recipes. By analyzing the hierarchical structure of ingredients, their approach captures partial overlaps and variations between recipes, emphasizing the importance of sub-component similarity in recipe matching.

\subsection{Domain Similarity}

Recent studies have extended recipe similarity beyond lexical and textual features to include domain-specific information, such as nutritional content. Ispirova et al.~\cite{13,14} proposed a framework for evaluating semantic and nutritional similarity between recipes. Semantic similarity was measured using transformer-based embeddings of cooking instructions, while domain expertise was combined with these semantic representations to learn feature embeddings that were subsequently used to predict nutritional attributes.

Ratisoontorn~\cite{15} introduced a recommendation method tailored for toddlers, which incorporated nutritional knowledge to determine recipe similarity. Their approach combined TF-IDF vectors of ingredients with domain-specific constraints, demonstrating the benefit of integrating nutritional information in similarity calculations.

\subsection{Semantic Similarity}

Zhu et al.~\cite{16} proposed methods for comparing recipes using fine-tuned transformer architectures to generate vector representations of ingredient sets. While effective in capturing ingredient-based similarity, this approach does not take into account cooking instructions, which are crucial to determining recipe similarity when ingredients are shared but cooking methods differ.

Salvador et al.~\cite{17} introduced a more comprehensive approach by leveraging transformer-based embeddings of recipe titles, ingredients, and instructions. This method captures semantic information across all textual components of a recipe, highlighting the importance of combining multiple sources of textual data for accurate similarity assessment.

\subsection{Data}

For our research, we used the Recipe1M dataset~\cite{18}, a comprehensive collection of recipes, each comprising a unique combination of ingredients and cooking instructions. The dataset enables detailed exploration of the nutritional value of dishes. Each recipe follows a standardized structure, including title, ID, ingredients, quantities, preparation instructions, and nutritional values. The ingredients are represented as a list in which each entry includes a main ingredient and its sub-characteristics. The instructions are organized sequentially to preserve the proper execution of the recipe. Nutritional information is provided in two forms: (i) per ingredient, covering fats, energy, protein, saturates, sodium, and sugars; and (ii) per 100 g of the recipe.

Despite the progress in recipe similarity research, several limitations remain. Existing approaches often focus on a single aspect of similarity, such as lexical overlap of ingredients, domain-specific nutritional features, or semantic embeddings of instructions. However, recipes are inherently multi-faceted, and similarity cannot be fully captured by considering any one perspective in isolation. Consequently, there is a need for a unified, multi-view approach that combines lexical, semantic, and nutritional information to provide a more all-rounded and accurate assessment of recipe similarity, which forms the basis of the method proposed in this study.

\section{Methods}

To address the challenges of recipe similarity detection, we propose a novel multi-view approach that fuses three perspectives: (i) semantic, (ii) lexical, and (iii) nutritional (domain knowledge). Each view leverages different input information from the recipe to compute a similarity score. Specifically, semantic similarity is measured using two transformer-based embedding methods; lexical similarity is computed using Jaccard similarity; and nutritional similarity is evaluated using two measures, one based on overall nutritional values and another based on nutritional similarity at the ingredient level.

In all methods, pairwise comparisons are performed by iterating through the recipes. At each step, one recipe is designated as the main recipe (the reference recipe), which is compared against all other recipes, referred to as secondary recipes. Each comparison creates a recipe pair containing the main recipe and the secondary recipe. At the end, we perform a weighted sum fusion of all three types of measures to get the final similarity score.

\subsection{Semantic Similarity through Transformer-Based Embeddings}

To measure the semantic similarity of recipes, we utilize the instruction data, which incorporates both the ingredients and the corresponding cooking methods. We use the fact that the ingredients are mentioned in the instructions so we do not add the ingredient separately. To represent the recipe instructions, sentence embeddings are employed. We benchmark two transformer models: DistilRoBERTa, chosen for its performance, and MiniLM-L6, selected for its faster computation time. For each recipe pair, embeddings are generated using both models, and their similarity is computed via cosine similarity. The resulting cosine similarities are normalized to fall between 0 and 1.

\begin{figure}[htbp]
    \centering
    \includegraphics[width=0.45\textwidth]{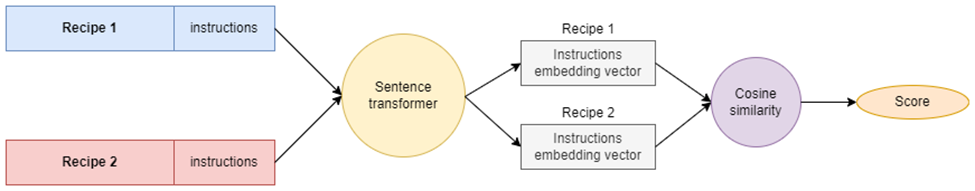}
    \caption{Graphical representation of the semantic similarity algorithm}
    \label{fig:semantic_algorithm}
\end{figure}

\subsection{Lexical Similarity through Ingredient Similarity}

To measure the similarity between recipe ingredients, we propose a modified version of the Jaccard similarity measure that leverages the hierarchical structure of the Recipe1M dataset. Ingredients in this dataset are represented hierarchically, progressing from general categories to specific types (e.g., "wheat flour, white, all-purpose, unenriched").

\begin{figure}[htbp]
\centerline{\includegraphics[width=0.40\textwidth]{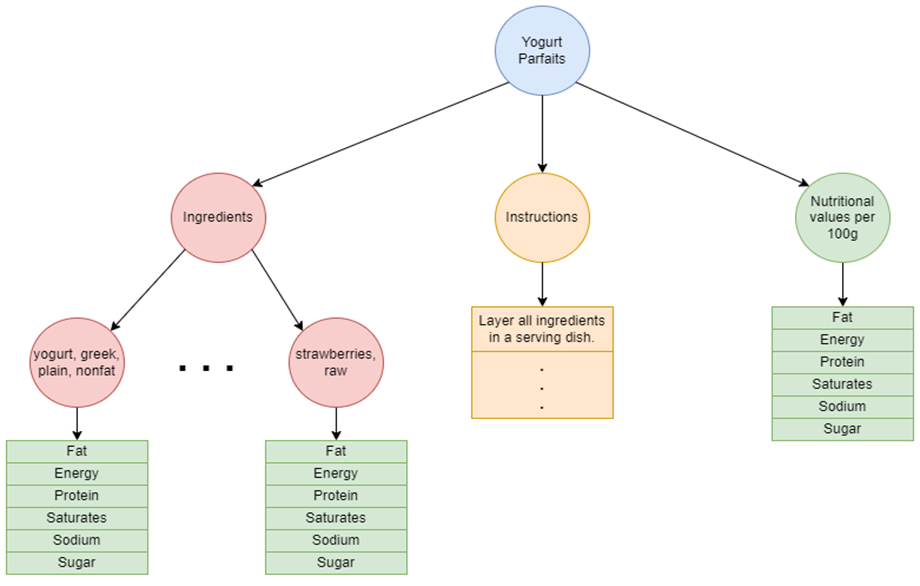}}
\caption{Example of a recipe from the Recipe1M dataset}
\label{fig:recipe_example}
\end{figure}

Our algorithm utilizes this structure by assigning a full similarity point when an entire ingredient from the main recipe appears in the secondary recipe, and a fractional score when only a more general subcategory matches. For example, if one recipe contains "spices, pepper, black" and another contains "spices, pepper, red or cayenne," two out of the three hierarchical components overlap, resulting in a similarity of 2/3 (0.66).

Next, we construct a similarity matrix comparing every ingredient in the main recipe with every ingredient in the secondary recipe. The Hungarian algorithm~\cite{21} is then applied to this matrix to determine the optimal one-to-one pairing of ingredients that maximizes the total similarity score. To account for recipes with different numbers of ingredients, we append null ingredients represented by zero vectors, allowing the algorithm to pair unmatched ingredients with these nulls.

Finally, the similarity scores of all matched ingredient pairs are averaged and normalized, ensuring the final similarity value lies within the [0, 1] range.

\begin{figure}[htbp]
    \centering
    \includegraphics[width=0.45\textwidth]{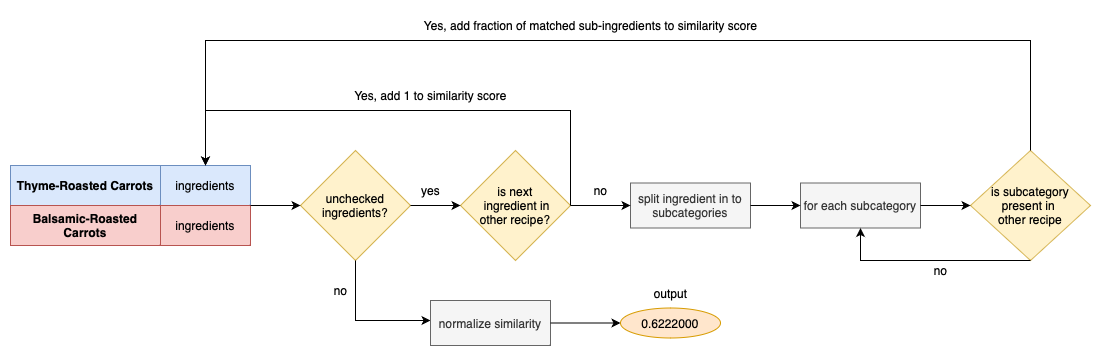}
    \caption{Visual representation of the algorithm used for calculating lexical similarity}
    \label{fig:lexical_similarity}
\end{figure}

\subsection{Domain Similarity}

\subsubsection{Nutritional Value per Recipe}

To compare the nutritional value of two recipes, we use the nutritional information associated with each recipe, including sugar, salt, saturates, energy, protein, and fat. This allows us to represent each recipe as a six-dimensional vector. We calculate the cosine similarity between each pair of recipes. The cosine similarity is normalized so it contains values between 0 and 1.

\begin{figure}[htbp]
    \centering
    \includegraphics[width=0.40\textwidth]{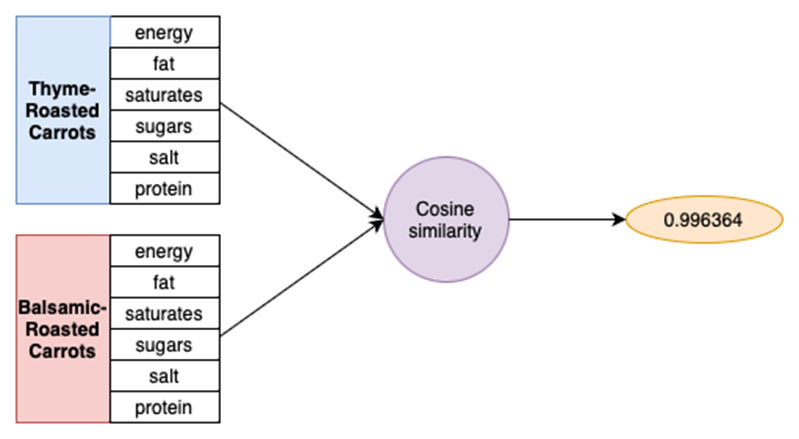}
    \caption{Graphical representation of the nutritional similarity between two recipes}
    \label{fig:nutritional_similarity_recipe}
\end{figure}

\subsubsection{Nutritional Value per Ingredient}

To fully utilize the available data, we calculate similarity based on the nutritional values per ingredient. We represent each ingredient's nutritional values as a vector.

For two recipes, we first standardize the nutrient vectors to reduce the effect of different scales. Then, we compute the cosine similarity between each ingredient in the first recipe and every ingredient in the second recipe, forming a similarity matrix.

To account for different numbers of ingredients and to maximize overall similarity, we find the optimal one-to-one matching between ingredients using the Hungarian algorithm. This produces a set of matched ingredient pairs with the highest total similarity. Finally, we compute the mean of these matched similarities, yielding a score between 0 and 1.

\begin{figure}[htbp]
    \centering
    \includegraphics[width=0.45\textwidth]{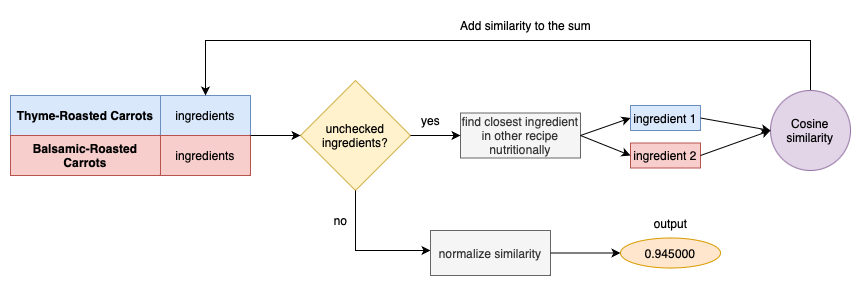}
    \caption{Visual representation of the algorithm used for nutritional similarity between recipes based on ingredients}
    \label{fig:nutritional_similarity_ingr}
\end{figure}

\subsection{Fusing the Similarity Methods}

To create a similarity metric, all the mentioned methods need to be combined. We summarize the methods with adjustable weights, which are initially set to 1/3 to have equal influence. The resulting similarity score ranges from 0 to 1, because all of the methods also range between 0 and 1. Since both semantic and nutritional similarity each involve two methods, we first compute the average of the two nutritional similarity methods and the average of the two semantic similarity methods before summing them.

\section{Analysis of Similarity Results}

\subsection{Analysis of Specific Examples}

This section presents an in-depth analysis of selected recipe pairs from our dataset, illustrating different scenarios of similarity as measured by semantic, lexical, and nutritional metrics. The cases range from high all-around similarity to clear non-matches and highlight interesting method behaviors.

\subsubsection{Case 1: High All-Around Similarity}

\textbf{Recipes:} Easy Lemonade and Lemon Granita

\begin{table}[htbp]
\centering
\caption{Similarity Scores for Case 1}
\begin{tabular}{| c | c |}
\hline
\textbf{Metric} & \textbf{Score} \\
\hline
Semantic (DistilRoBERTa) & 0.9015 \\
\hline
Semantic (MiniLM-L6) & 0.8720 \\
\hline
Lexical & 1.0000 \\
\hline
Nutritional (per recipe) & 1.0000 \\
\hline
Nutritional (per ingredients) & 1.0000 \\
\hline
\end{tabular}
\label{tab:case1_scores}
\end{table}

\textbf{Analysis:} A clear match. As shown in Table \ref{tab:case1_scores}, the Lexical and Nutritional scores are both 1.0000. The recipes are lexically identical with the same three ingredients. The instructions, while slightly different (pan vs. saucepan, melts vs. complete dissolve), describe the exact same process (make a simple syrup, add lemon), resulting in high semantic similarity. The nutritional profiles are also nearly identical, confirming the overall match.

\textbf{Supporting Data:}
\begin{itemize}
\item Ingredients A: lemon juice, raw; water, bottled, generic; sugars, granulated
\item Ingredients B: water, bottled, generic; sugars, granulated; lemon juice, raw
\item Nutritional Vector A: [0.02, 0.03, 0.01, 0.0, 4.83]
\item Nutritional Vector B: [0.03, 0.05, 0.01, 0.01, 15.45]
\item Instructions A: In a large pan, combine water and sugar. Heat until the sugar just melts. Remove from heat and pour...
\item Instructions B: Pour the water and sugar into a small saucepan, and bring to a boil. Boil until the sugar is completely dissolved...
\end{itemize}

\subsubsection{Case 2: High Semantic, Low Lexical \& Nutritional}

\textbf{Recipes:} French Dressing and The Texican Cocktail

\begin{table}[htbp]
\centering
\caption{Similarity Scores for Case 2}
\begin{tabular}{| c | c |}
\hline
\textbf{Metric} & \textbf{Score} \\
\hline
Semantic (DistilRoBERTa) & 0.8689 \\
\hline
Semantic (MiniLM-L6) & 0.7803 \\
\hline
Lexical & 0.0091 \\
\hline
Nutritional (per recipe) & 0.0407 \\
\hline
Nutritional (per ingredients) & 0.9271 \\
\hline
\hline
\end{tabular}
\label{tab:case2_scores}
\end{table}

\textbf{Analysis:} This pair highlights the ``process-driven'' nature of semantic similarity. As detailed in Table \ref{tab:case2_scores}, both instructions involve combining ingredients and shaking, which semantic models correctly identify as similar (scores $\ge 0.78$). However, the recipes themselves are very different (oil dressing vs. tequila cocktail), as shown by the near zero lexical score (0.0091) and extremely low recipe-level nutritional score (0.0407).

\textbf{Supporting Data:}
\begin{itemize}
\item Ingredients A: oil, olive, salad or cooking; vinegar, red wine; water, bottled, generic; sugars, granulated; mustard, prepared, yellow; salt, table; sauce, worcestershire; spices, paprika
\item Ingredients B: alcoholic beverage, tequila sunrise, canned; cranberry juice, unsweetened; lime juice, raw
\item Nutritional Vector A: [6.03, 0.08, 0.19, 0.83, 0.15]
\item Nutritional Vector B: [0.12, 0.36, 0.06, 0.01, 7.22]
\item Instructions A: Combine all ingredients in a jar. Shake well...
\item Instructions B: In a cocktail shaker with ice, shake all ingredients, strain into a chilled glass...
\end{itemize}

\subsubsection{Case 3: High Lexical, Low Semantic (Not Found)}

No examples were found for this case. Pairs with high lexical similarity (above 0.7) but low semantic similarity (less than 0.6) do not appear in the dataset. This suggests that recipes with highly overlapping ingredients also tend to have similar instructions. We also need to take into consideration that only a subset of 410 recipes was used and even though the resulting recipe pairs represent a substantial amount, not every edge case can be captured.

\subsubsection{Case 4: High Nutritional (Recipe), Low Semantic \& Lexical}

\textbf{Recipes:} Bean Jam (Anko) and Metropolitan Martini

\begin{table}[htbp]
\centering
\caption{Similarity Scores for Case 4}
\begin{tabular}{| c | c |}
\hline
\textbf{Metric} & \textbf{Score} \\
\hline
Semantic (DistilRoBERTa) & 0.5086 \\
\hline
Semantic (MiniLM-L6) & 0.6114 \\
\hline
Lexical & 0.0000 \\
\hline
Nutritional (per recipe) & 0.9997 \\
\hline
Nutritional (per ingredient) & 0.5639 \\
\hline
\end{tabular}
\label{tab:case4_scores}
\end{table}

\textbf{Analysis:} This is an example of a ``coincidental match'' showing the limitations of relying on a single nutritional vector. As shown in Table \ref{tab:case4_scores}, the recipes themselves have nothing in common (Lexical score 0.0000), but their 5-nutrient macro profiles are coincidentally similar, leading to an almost perfect Nutritional score per recipe of 0.9997. The low semantic scores (around 0.50-0.61) correctly reject the match, emphasizing the ensemble's ability to filter these false positives.

\textbf{Supporting Data:}
\begin{itemize}
\item Ingredients A: beans, snap, green, raw; sugars, granulated; salt, table
\item Ingredients B: alcoholic beverage, distilled, gin, 90 proof; cranberry juice, unsweetened; lime juice, raw
\item Nutritional Vector A: [0.12, 1.02, 0.09, 0.03, 46.13]
\item Nutritional Vector B: [0.07, 0.23, 0.01, 0.01, 5.39]
\item Instructions A: Wash the beans, put them in a pressure cooker pot (or just a pot), then pour in water until they...
\item Instructions B: In a cocktail shaker filled halfway with ice combine all ingredients and shake well. Strain mixture...
\end{itemize}

\subsubsection{Case 5: Low All-Around Similarity (Clear Non-Match)}

\textbf{Recipes:} Roasted Potatoes and The Texican Cocktail

\begin{table}[htbp]
\centering
\caption{Similarity Scores for Case 5}
\begin{tabular}{| c | c |}
\hline
\textbf{Metric} & \textbf{Score} \\
\hline
Semantic (DistilRoBERTa) & 0.4875 \\
\hline
Semantic (MiniLM-L6) & 0.4761 \\
\hline
Lexical & 0.0000 \\
\hline
Nutritional (per recipe) & 0.0442 \\
\hline
Nutritional (per ingredient) & 0.2740 \\
\hline
\end{tabular}
\label{tab:case5_scores}
\end{table}

\textbf{Analysis:} A clear non-match. As expected for dissimilar recipes (see Table \ref{tab:case5_scores}), this pair shares no ingredients, cooking methods, or nutritional profile. All scores are correctly low (Lexical 0.0000, Semantic $\approx$ 0.48, Nutritional $\approx$ 0.04), representing a baseline for non-similar recipes.

\textbf{Supporting Data:}
\begin{itemize}
\item Ingredients A: potatoes, raw, skin; oil, olive, salad or cooking; salt, table
\item Ingredients B: alcoholic beverage, tequila sunrise, canned; cranberry juice, unsweetened; lime juice, raw
\item Nutritional Vector A: [2.97, 2.48, 0.49, 0.42, 0.0]
\item Nutritional Vector B: [0.12, 0.36, 0.06, 0.01, 7.22]
\item Instructions A: Preheat oven to 400F. Toss potatoes with oil and sea salt in a large baking pan, then arrange them...
\item Instructions B: In a cocktail shaker with ice, shake all ingredients, strain into a chilled glass...
\end{itemize}

\subsubsection{Case 6: Semantic Model Disagreement}

\textbf{Recipes:} Taco Seasoning and Freezer Apple Pie Filling - OAMC

\begin{table}[htbp]
\centering
\caption{Similarity Scores for Case 6}
\begin{tabular}{| c | c |}
\hline
\textbf{Metric} & \textbf{Score} \\
\hline
Semantic (DistilRoBERTa) & 0.5200 \\
\hline
Semantic (MiniLM-L6) & 0.7127 \\
\hline
Lexical & 0.1125 \\
\hline
Nutritional (per recipe) & 0.3892 \\
\hline
Nutritional (per ingredient) & 0.6725 \\
\hline
\end{tabular}
\label{tab:case6_scores}
\end{table}

\textbf{Analysis:} This case demonstrates the value of benchmarking multiple semantic models. As highlighted in Table \ref{tab:case6_scores}, MiniLM-L6 identifies a moderately high similarity (0.7127), while DistilRoBERTa finds a lower similarity (0.5200). This disagreement is likely because both recipes involve general instructions like ``mixing dry ingredients'' (spices vs. apples/sugar/cornstarch). MiniLM-L6 may be focusing on this general process, while the more robust DistilRoBERTa model correctly identifies that a spice mix and an apple pie filling are not truly similar. The ensemble method helps to mitigate the effects of such model-specific overestimation.

\textbf{Supporting Data:}
\begin{itemize}
\item Ingredients A: spices, chili powder; spices, garlic powder; spices, onion powder; spices, pepper, red or cayenne; spices, oregano, dried; spices, paprika; spices, cumin seed; salt, table; spices, pepper, black
\item Ingredients B: apples, raw, with skin; lemon juice, raw; sugars, granulated; cornstarch; spices, cinnamon, ground; spices, nutmeg, ground; salt, table; water, bottled, generic
\item Nutritional Vector A: [6.74, 12.74, 4.06, 1.28, 6.19]
\item Nutritional Vector B: [0.09, 0.13, 0.12, 0.02, 20.66]
\item Instructions A: In a small bowl, mix together chili powder, garlic powder, onion powder, red pepper flakes...
\item Instructions B: In a large bowl, toss apples with lemon juice and set aside. Pour water into a Dutch oven over medium...
\end{itemize}

Across these cases, the analysis demonstrates the strengths and limitations of the different similarity metrics we presented. High semantic, lexical, and nutritional agreement typically corresponds to clear matches, while low scores indicate non-matches. Some cases highlight model disagreements or coincidental matches, emphasizing the importance of considering multiple metrics together when assessing recipe similarity.

\subsection{Score Distribution: The ``Personality'' of Each Metric}

Table \ref{tab:metric_stats} presents descriptive statistics for the main similarity metrics across the entire dataset. This analysis reveals the characteristic behavior of each metric.

\begin{table}[htbp]
\centering
\caption{Descriptive Statistics of Core Similarity Metrics}
\resizebox{\columnwidth}{!}{%
\begin{tabular}{|l|c|c|c|c|c|c|}
\hline
\textbf{Metric} & \textbf{Mean} & \textbf{Median} & \textbf{Std. Dev.} & \textbf{Skew} & \textbf{Min} & \textbf{Max} \\
\hline
DistilRoBERTa (Semantic) & 0.6789 & 0.6739 & 0.0734 & 0.7216 & 0.4327 & 1.0000 \\
\hline
MiniLM-L6 (Semantic) & 0.6971 & 0.6930 & 0.0683 & 0.6854 & 0.4727 & 1.0000 \\
\hline
Nutritional (Recipe) & 0.6339 & 0.6893 & 0.2929 & -0.4831 & 0.0000 & 1.0000 \\
\hline
Ingredients Jaccard (Lexical) & 0.1072 & 0.0769 & 0.1391 & 2.4875 & 0.0000 & 1.0000 \\
\hline
Ingredients Nutritional & 0.7733 & 0.8000 & 0.1661 & -0.8777 & 0.0000 & 1.0000 \\
\hline
\end{tabular}%
}
\label{tab:metric_stats}
\end{table}

The Lexical metric is heavily positively skewed ($\text{Skew}=2.49$) with a low mean ($\approx 0.11$). This confirms that most recipe pairs in the dataset are lexically unique, highlighting the necessity of complementary non-lexical views to measure conceptual similarity.

Semantic metrics (DistilRoBERTa, MiniLM-L6) are tightly clustered (Std. Dev. $\approx 0.07$) with high means ($\approx 0.68$). This suggests they find a baseline level of moderate similarity for nearly all recipes due to shared language context, making them less discriminative for identifying precise matches.

The Nutritional (Recipe) metric shows a high median (0.6893) but the largest dispersion (Std. Dev. 0.2929). This broad spread across the score range confirms its high susceptibility to both false positives and false negatives.

\begin{figure}[htbp]
    \centering
    \includegraphics[width=0.45\textwidth]{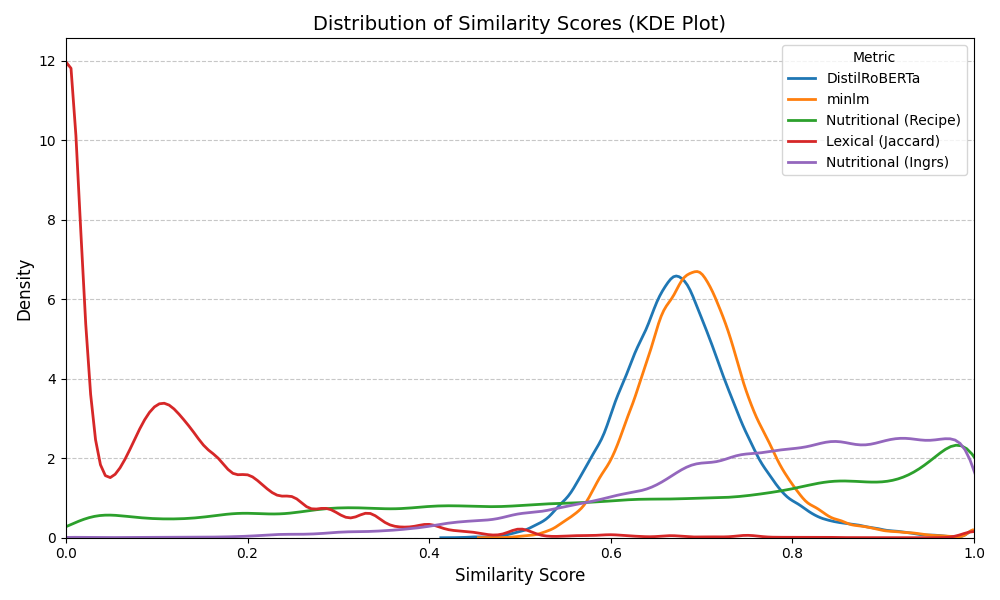}
    \caption{Distribution of Similarity Scores (KDE Plot)}
    \label{fig:similarity_distribution_core_kde}
\end{figure}

\subsection{Correlation Analysis: Checking Redundancy}

Table \ref{tab:metric_corr} shows the correlation between the main similarity views, demonstrating their relative independence.

\begin{table*}
\centering
\caption{Correlation Heatmap of Similarity Metrics}
\begin{tabular}{|l|c|c|c|c|c|}
\hline
& DistilRoBERTa & MiniLM-L6 & Nutritional (Recipe) & Jaccard & Nutritional (Ingredient)\\
\hline
DistilRoBERTa & 1.00 & 0.85 & 0.25 & 0.54 & 0.21 \\
\hline
MiniLM-L6 & 0.85 & 1.00 & 0.26 & 0.52 & 0.22 \\
\hline
Nutritional & 0.25 & 0.26 & 1.00 & 0.23 & 0.26 \\
\hline
Jaccard & 0.54 & 0.52 & 0.23 & 1.00 & 0.39 \\
\hline
Ingredients Nutritional & 0.21 & 0.22 & 0.26 & 0.39 & 1.00 \\
\hline
\end{tabular}
\label{tab:metric_corr}
\end{table*}

The correlation data strongly validates the multi-view approach by establishing three independent signals:

\begin{itemize}
\item The two semantic models (DistilRoBERTa and MiniLM-L6) exhibit a high correlation ($\mathbf{r=0.85}$), affirming their shared role as a unified Semantic View.
\item The Semantic View shows a low correlation with the Nutritional View ($\mathbf{r \approx 0.25}$) and only a moderate correlation with the Lexical View ($\mathbf{r \approx 0.54}$).
\item The Lexical and Nutritional views exhibit a very low correlation ($\mathbf{r=0.23}$).
\end{itemize}

These low cross-correlations prove that the three primary views---Semantic, Lexical, and Nutritional---are largely non-redundant and capture unique, distinct dimensions of recipe similarity.

\begin{figure}[htbp]
    \centering
    \includegraphics[width=0.40\textwidth]{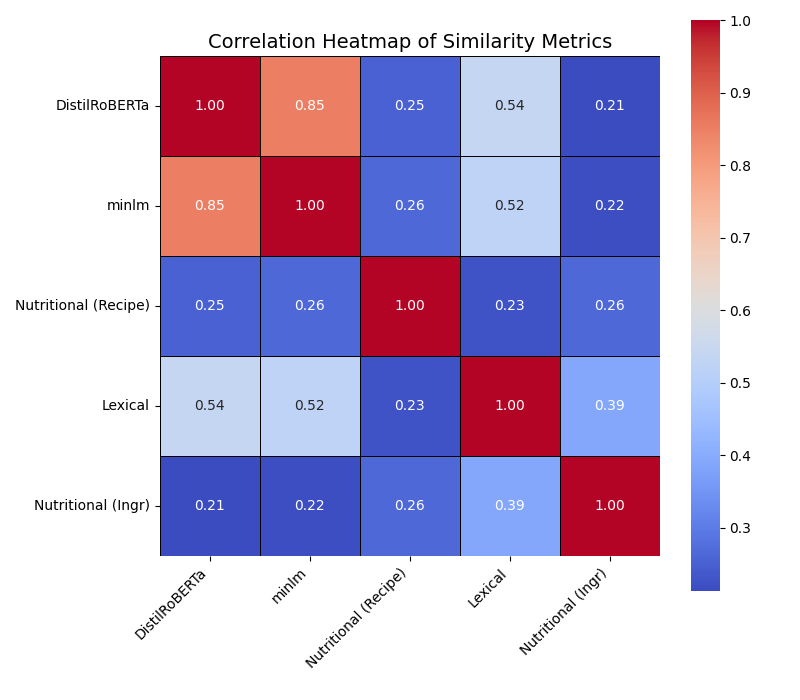}
    \caption{Correlation Heatmap of Similarity Metrics}
    \label{fig:correlation_heatmap_core}
\end{figure}

\subsection{Failure Case Analysis}

Table \ref{tab:failure_cases} quantifies instances where a single metric is misleading relative to the other views.

\begin{table}[htbp]
\centering
\caption{Failure Cases Across Metrics}
\resizebox{\columnwidth}{!}{%
\begin{tabular}{|l|l|c|c|}
\hline
\textbf{Metric} & \textbf{Criteria} & \textbf{Count (N)} & \textbf{Percentage} \\
\hline
Nutritional & $\text{nutr} > 0.95$, $\text{roberta} < 0.6$, $\text{jaccard} < 0.1$ & 1,738 & 1.03\% \\
\hline
Semantic & $\text{roberta} > 0.85$, $\text{jaccard} < 0.1$, $\text{nutr} < 0.2$ & 12 & 0.01\% \\
\hline
Lexical & $\text{jaccard} > 0.3$, $\text{roberta} < 0.6$ & 89 & 0.05\% \\
\hline
\end{tabular}%
}
\label{tab:failure_cases}
\end{table}

The high count of 1,738 nutritional failures is a direct consequence of the Nutritional metric's broad distribution (Table \ref{tab:metric_stats}). This establishes that nutritional similarity alone is the primary source of false positives in the dataset. The 12 semantic failures, though rare, confirm that even high-performing language models can overgeneralize similarity based on common instruction phrasing, failing to capture the unique ingredient and nutritional context. The 89 lexical failures show that ingredient overlap is not sufficient to guarantee semantic similarity, as identical ingredients can be used for entirely different dishes. Collectively, these failures prove that no single metric can achieve both high precision (avoiding false positives) and high recall (avoiding false negatives) across all similarity dimensions.

\subsection{Agreement Analysis: Ensemble Validation}

Table \ref{tab:ensemble_agreement} shows how lexical similarity drives the ensemble into agreement with semantic and nutritional metrics.

\begin{table}[htbp]
\centering
\caption{Average Metric Scores by Lexical Jaccard Bins}
\resizebox{\columnwidth}{!}{%
\begin{tabular}{|c|c|c|c|}
\hline
\textbf{Lexical Bin} & \textbf{Avg. Semantic} & \textbf{Avg. Nutritional} & \textbf{Final Ensemble Score} \\
\hline
0.0–0.1 & 0.66 & 0.60 & 0.45 \\
\hline
0.1–0.2 & 0.68 & 0.64 & 0.52 \\
\hline
0.2–0.3 & 0.71 & 0.68 & 0.57 \\
\hline
0.3–0.4 & 0.74 & 0.75 & 0.64 \\
\hline
0.4–0.5 & 0.78 & 0.86 & 0.70 \\
\hline
0.5–0.6 & 0.81 & 0.91 & 0.75 \\
\hline
0.6–0.7 & 0.84 & 0.93 & 0.81 \\
\hline
0.7–0.8 & 0.82 & 0.97 & 0.85 \\
\hline
0.8–0.9 & 0.89 & 0.94 & 0.89 \\
\hline
0.9–1.0 & 0.88 & 0.93 & 0.92 \\
\hline
\end{tabular}%
}
\label{tab:ensemble_agreement}
\end{table}

The upward trend across all metrics validates the ensemble's approach. The Final Ensemble Score exhibits the most stable and reliable increase, moving from 0.45 for low-Jaccard pairs to 0.92 for near-duplicates, demonstrating its successful fusion of signals. The ensemble's filtering capability is visible in the low Jaccard bins. For instance, in the 0.0-0.1 bin, the Nutritional score is 0.60, but the final ensemble score is successfully pulled down to 0.45, mitigating the nutritional false positives identified in Table \ref{tab:failure_cases}. The ensemble provides a coherent, validated measure that consistently reflects true, multi-dimensional recipe proximity, proving its superior reliability over any single component.

\subsection{Semantic Model Comparison}

Table \ref{tab:semantic_comparison} compares the performance of DistilRoBERTa and MiniLM-L6 semantic models.

\begin{table}[htbp]
\centering
\caption{DistilRoBERTa vs. MiniLM-L6 Semantic Scores}
\begin{tabular}{|l|c|}
\hline
\textbf{Statistic} & \textbf{Value} \\
\hline
Mean Absolute Difference & 0.0344 \\
\hline
Max Absolute Difference & 0.1927 \\
\hline
Correlation & 0.85 \\
\hline
\end{tabular}
\label{tab:semantic_comparison}
\end{table}

The high correlation ($\mathbf{r=0.85}$) and small mean absolute difference ($\mathbf{0.0344}$) demonstrate that the computationally efficient MiniLM-L6 model provides a high-fidelity score nearly identical to DistilRoBERTa. This justifies the efficiency gains. By averaging the two scores, we leverage the high agreement while ensuring that the few cases of outlier disagreement are stabilized, leading to a more robust semantic view for the final ensemble.

\begin{figure}[htbp]
    \centering
    \includegraphics[width=0.40\textwidth]{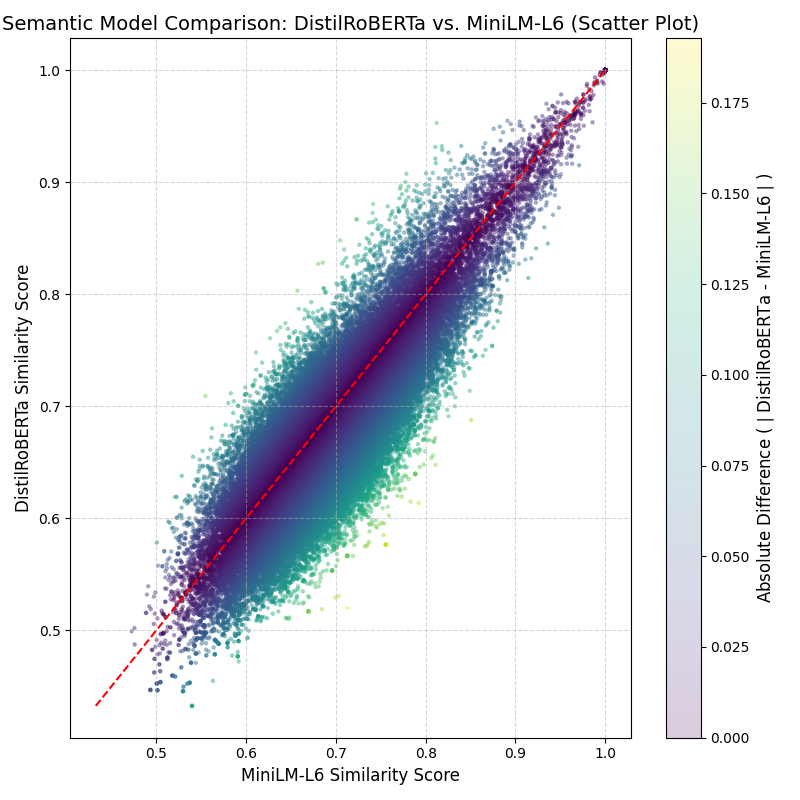}
    \caption{Semantic Model Comparison: DistilRoBERTa vs. MiniLM-L6 (Scatter Plot)}
    \label{fig:semantic_model_comparison_core_scatter}
\end{figure}

\section{Evaluation}

\subsection{Human-Computer Interaction Tool}

Due to the absence of a gold standard for evaluation, a web application was developed specifically to allow domain experts to assess the effectiveness of algorithmically generated results. The application presents a set of 100 unique recipes, and beneath each recipe, it displays one or more of the most similar recipes. The similar recipes displayed are determined by selecting the top 20\% most similar recipes for each entry.

Out of a total of 318 recipe pairs evaluated by two domain experts, both experts agreed on 255 pairings (80\%), while their assessments differed for the remaining 63 (20\%).

\begin{figure}[htbp]
    \centering
    \includegraphics[width=0.45\textwidth]{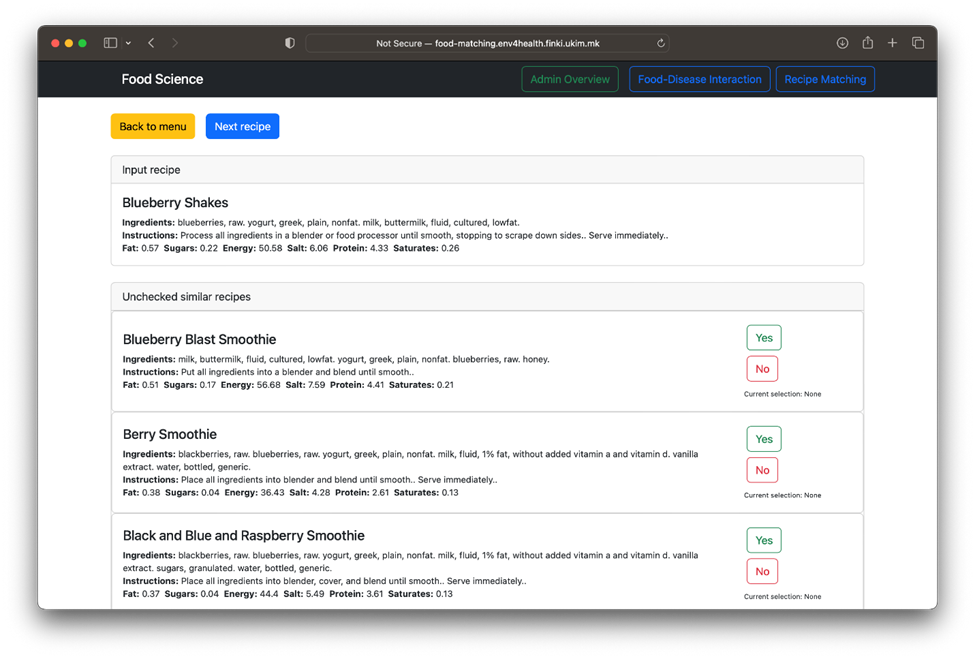}
    \caption{Example web-based tool for evaluating the results}
    \label{fig:website_example}
\end{figure}

This domain expert validation allows us to create a ground truth annotated dataset. We later use this dataset to train two models, Logistic Regression and Random Forest, meaning that their learned feature importance reflect the implicit reasoning of human evaluators.

\subsection{Logistic Regression}

Logistic regression is a statistical model used for binary classification. The model was trained on a set of features---including semantic, nutritional, and lexical features---and its predictions were evaluated against the results from the evaluation, specifically on the 255 recipe pairs that the experts agreed upon, using 5-fold cross-validation.

The logistic regression model achieved an average accuracy of \textbf{89\%} with a standard deviation of \textbf{5\%}. As shown in Table \ref{tab:classification_report_lr}, Class 0 (``not similar'') achieved high precision (0.90) and recall (0.95), suggesting that most non-similar recipe pairs were correctly identified with minimal false positives or false negatives. Class 1 (``similar'') achieved a precision of 0.86 and recall of 0.75, indicating that the model correctly identified most true cases of similarity but missed some true positives, likely due to class imbalance.

Overall, these metrics indicate that logistic regression performs reasonably well, with relatively high overall accuracy. However, the model performs better on the majority class (not similar), likely due to class imbalance.

\begin{table}[htbp]
\caption{Logistic Regression Model Classification Report}
\centering
\begin{tabular}{|c|c|c|c|}
\hline
\textbf{Class} & \textbf{Precision} & \textbf{Recall} & \textbf{F1-score} \\
\hline
0 (Not Similar) & 0.90 & 0.95 & 0.92 \\
\hline
1 (Similar) & 0.86 & 0.75 & 0.80 \\
\hline
\end{tabular}
\label{tab:classification_report_lr}
\end{table}

In logistic regression, the coefficients represent the change in the log-odds of the target variable for a one-unit change in each feature. A positive coefficient increases the probability of the positive class (1), while a negative coefficient decreases it.

\begin{table}[htbp]
\caption{Logistic Regression Coefficients and Normalized Importance}
\centering
\begin{tabular}{|c|c|c|}
\hline
\textbf{Feature} & \textbf{Coefficient} & \textbf{Normalized Importance} \\
\hline
Semantic (Preparation steps) & 0.087156 & 4.1\% \\
\hline
Domain (Nutritional values) & 0.063678 & 3.0\% \\
\hline
Lexical (Ingredients / Jaccard) & 1.986991 & 92.9\% \\
\hline
\end{tabular}
\label{tab:logreg_coefficients}
\end{table}

\subsubsection{Interpretation of the Logistic Regression Coefficients}

As shown in Table \ref{tab:logreg_coefficients}:

\textbf{Lexical:} The large positive coefficient indicates that as ingredient overlap increases (higher Jaccard index), the probability that a recipe pair is labeled as ``similar'' (1) also increases. This strong influence highlights lexical similarity as the dominant predictor of recipe similarity, consistent with the Random Forest model results.

\textbf{Semantic:} The small positive coefficient suggests that higher semantic similarity between preparation steps slightly increases the probability of being classified as ``similar.'' While this contribution is modest, it still positively affects prediction.

\textbf{Domain:} Nutritional similarity has a very small positive effect, suggesting a minimal influence on the model's prediction. This aligns with the Random Forest model, where nutritional features contributed less than lexical similarity.

\subsection{Random Forest}

The Random Forest model is an ensemble method that builds multiple decision trees and combines their results to improve classification accuracy. Each tree makes its own prediction, and the majority vote is used for the final output. This model is evaluated on the same 255 pairs on which the experts agreed.

The Random Forest model achieved an accuracy of \textbf{89\%} ± 0.02, comparable to that of the Logistic Regression model. As shown in Table \ref{tab:rf_report}, Class 0 (``not similar'') achieved precision and recall of 0.92, demonstrating strong performance in identifying non-similar recipes. Class 1 (``similar'') achieved precision 0.80 and recall 0.81, showing slightly lower but still high performance in correctly identifying similar recipe pairs. These results suggest that, while Random Forest is generally effective, it handles Class 1 predictions with slightly less accuracy, likely due to class imbalance, similar to Logistic Regression.

\begin{table}[htbp]
\caption{Random Forest Classification Report}
\centering
\begin{tabular}{|c|c|c|c|}
\hline
\textbf{Class} & \textbf{Precision} & \textbf{Recall} & \textbf{F1-score} \\
\hline
0 (Not Similar) & 0.89 & 0.91 & 0.90 \\
\hline
1 (Similar) & 0.76 & 0.72 & 0.74 \\
\hline
\end{tabular}
\label{tab:rf_report}
\end{table}

In the Random Forest model, feature importance is based on how much each feature contributes to the overall accuracy of the model. This is typically measured by the extent to which each feature helps reduce impurity (e.g., Gini impurity) in the decision trees within the forest.

\begin{table}[htbp]
\caption{Feature Importance in Random Forest Model (Normalized Embeddings)}
\centering
\begin{tabular}{|c|c|}
\hline
\textbf{Feature} & \textbf{Normalized Importance} \\
\hline
Lexical (Ingredients / Jaccard) & 42.6\% \\
\hline
Domain (Nutritional Values) & 37.3\% \\
\hline
Semantic (Preparation Steps) & 20.0\% \\
\hline
\end{tabular}
\label{tab:rf_importance}
\end{table}

\subsubsection{Interpretation of Feature Importance in the Random Forest Model}

As shown in Table \ref{tab:rf_importance}:

\textbf{Lexical:} This feature has the highest importance in the model (42.6\%), indicating that ingredient overlap (how similar the ingredients are between two recipes) is the most influential factor in determining recipe similarity. The high value demonstrates that lexical similarity strongly drives the model's predictions, highlighting the dominant role of ingredients when assessing recipe similarity.

\textbf{Domain:} Nutritional values (e.g., energy, proteins, fats, etc.) contribute the second highest importance (37.3\%). This shows that domain-specific information still plays a significant role in predicting recipe similarity, although it is less influential than lexical similarity.

\textbf{Semantic:} Semantic similarity of preparation steps has the lowest importance (20.0\%) among the three features. While it contributes to determining similarity, its impact is considerably smaller than that of lexical or nutritional features.

\begin{figure}[htbp]
    \centering
    \includegraphics[width=0.45\textwidth]{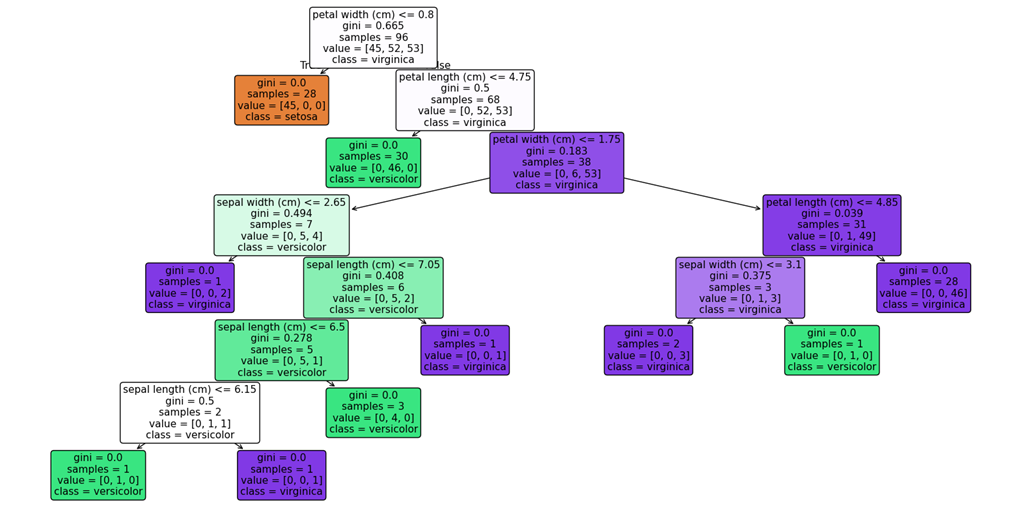}
    \caption{The constructed Random Forest model}
    \label{fig:random_forest_model}
\end{figure}

\section{Conclusion}

This study presents a comprehensive approach to measuring recipe similarity, combining semantic, lexical, and nutritional aspects. The results demonstrate that integrating these perspectives provides a more complete understanding of relationships between recipes. Semantic similarity effectively captures nuances in preparation instructions. Lexical similarity, computed using a modified Jaccard index that accounts for the hierarchical structure of ingredients, has a strong influence in traditional models. Nutritional similarity adds a domain-specific dimension, although its influence is moderate compared to the other features.

Both the Logistic Regression and Random Forest models achieved strong performance (89\% accuracy). The models revealed distinct feature dynamics. Logistic Regression showed that lexical similarity was by far the most influential predictor, while semantic and nutritional features contributed minimally. In contrast, the Random Forest model captured more complex, non-linear relationships, with all three feature types playing meaningful roles.

The evaluation results indicate that ingredient overlap alone is insufficient for accurately assessing similarity from a human perspective. A hybrid approach that combines lexical, semantic, and nutritional information is recommended. Dynamic weighting in a multi-perspective approach can further refine results for specific use cases. Future research may explore advanced model architectures and larger datasets to improve robustness. Additionally, investigating user-specific similarity metrics and analyzing ingredients based on their molecular structures could enhance personalization in recipe recommendation systems.

\end{document}